\documentclass[sigconf,nonacm]{acmart}
\AtBeginDocument{%
  }

\usepackage{graphicx}
\usepackage{enumitem}
\usepackage{listings}
\usepackage{pdflscape}
\usepackage[most]{tcolorbox}

\PassOptionsToPackage{table}{xcolor}
\usepackage{xcolor}
\usepackage[table]{xcolor}
\usepackage{siunitx}
\usepackage{multirow}
\begin{document}

\title{Customer-Agent: Overcoming Context Limitations in Ultra-Long Shopping Trajectories via Tool-Augmented Agents and RLVR}

\author{Hongye Liu}
\authornote{Work done during an internship at Amazon.}
\affiliation{%
  \institution{Duke University}
  \city{Durham}
  \country{USA}
}
\email{hongye.liu@duke.edu}

\author{Rongmei Lin}
\affiliation{%
  \institution{Amazon}
  \city{Seattle}
  \country{USA}}
\email{linrongm@amazon.com}

\author{Zhongruo Wang}
\affiliation{%
  \institution{Amazon}
  \city{Palo Alto}
  \country{USA}}
\email{ysxd@amazon.com}

\author{Anurag Kashyap}
\affiliation{%
  \institution{Amazon}
  \city{Seattle}
  \country{USA}}
\email{anurakas@amazon.com}

\author{Hejie Cui}
\affiliation{%
  \institution{Google}
  \city{Mountain View}
  \country{USA}}
\email{hcui24@cs.stanford.edu}

\author{Ricardo Henao}
\affiliation{%
  \institution{Duke University}
  \city{Durham}
  \country{USA}
}
\email{ricardo.henao@duke.edu}

\author{Besnik Fetahu}
\affiliation{%
  \institution{Amazon}
  \city{Seattle}
  \country{USA}}
\email{besnikf@amazon.com}

\author{Bing Yin}
\affiliation{%
  \institution{Amazon}
  \city{Seattle}
  \country{USA}}
\email{qingyy@amazon.com}


\begin{abstract}
Understanding customer shopping trajectories is essential for enabling personalized shopping experiences.
However, shopping records ({\em i.e.}, customer's search, clicks, purchases, {\em etc.}) often span long time horizons over multiple years, resulting in extremely long trajectories that pose significant challenges for existing large language models (LLMs). 
Despite the importance of this problem, existing benchmarks are limited to short customer trajectories, while real-world trajectories from large e-commerce platforms are rarely accessible due to data privacy constraints. 
To address this gap, we introduce ShopTrajQA, a long-context resource for shopping-trajectory reasoning constructed from real-world product information and simulated shopping trajectories.
ShopTrajQA comprises both evaluation and training data: the evaluation benchmark includes variants of up to 32k and 64k tokens, enabling a systematic evaluation of the robustness of the model under varying context lengths, while the training data consists of a supervised fine-tuning (SFT) set of verified multi-turn tool-use trajectories and a Reinforcement Learning with Verifiable Rewards (RLVR) set with executable, automatically-checkable answers.
Together, these subsets provide a complete resource for building and evaluating agents that reason over long shopping trajectories.
Through comprehensive benchmarking of 9 frontier LLMs, we identify critical performance gaps in reasoning over long shopping trajectory data.
To address these challenges, we propose a {\em Customer Agent Framework} for ultra-long context management.
Leveraging an RLVR agentic training paradigm, our approach stores trajectories as external local files and trains the agent to autonomously retrieve and parse them through code-interpreter interactions ({\em e.g.}, SQL queries), effectively bypassing the fixed in-context window constraints of LLMs.
Experimental results demonstrate that our framework achieves strong performance for ShopTrajQA and shows generalization to other complex reasoning tasks.
The project page is available at \url{https://hyfred.github.io/Customer-Agent}.
\end{abstract}





\maketitle

\begin{figure*}[t]
    \centering
    \includegraphics[width=1.9\columnwidth]{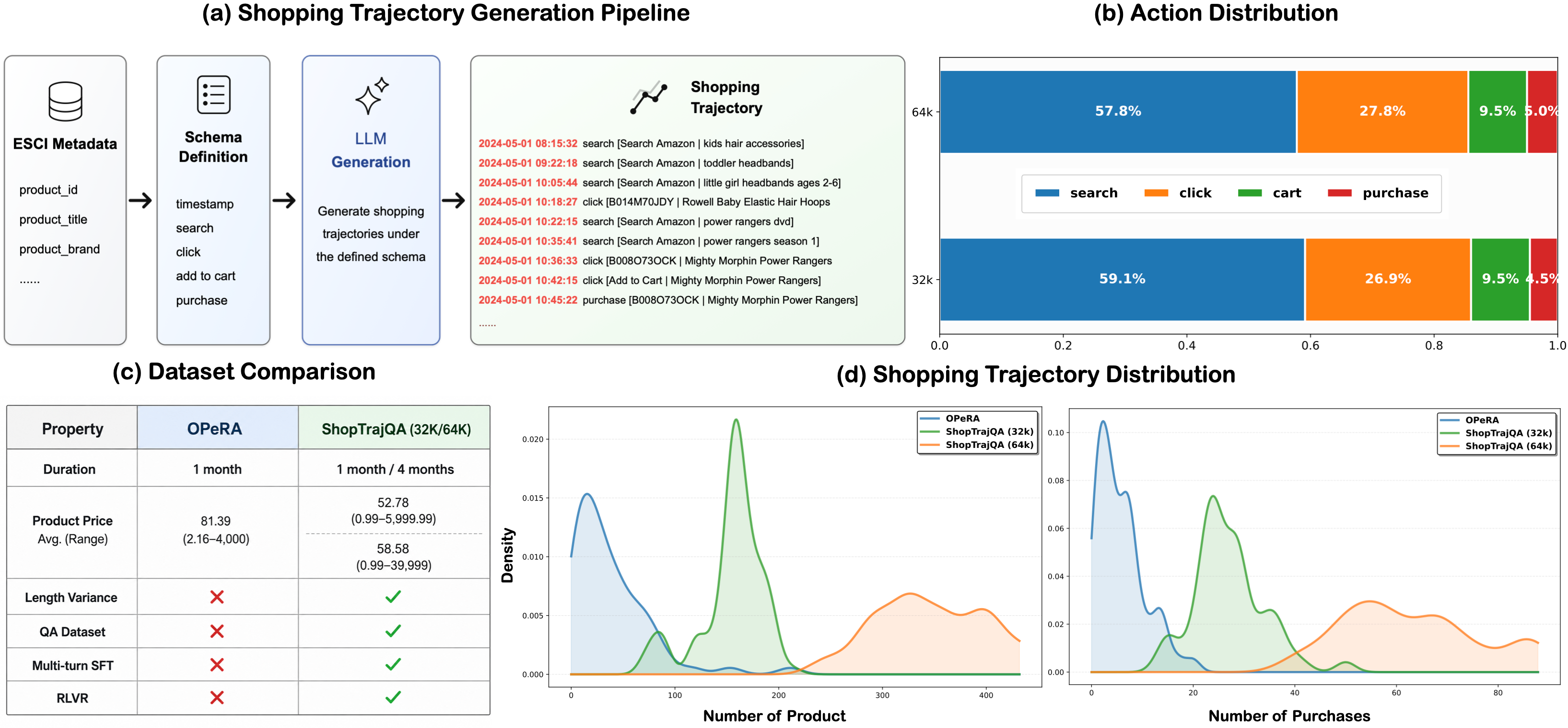}
    \caption{Comparison between ShopTrajQA and OPeRA. The first row illustrates the trajectory generation pipeline and the distribution of user actions. The second row presents a comparison of ShopTrajQA and OPeRA, along with the distributions of the number of products and purchases in each trajectory.}
    \vspace{-4mm}
    \label{fig:shop_data_construction}
\end{figure*}

\section{Introduction}
Understanding customer shopping trajectories is essential for providing high-quality and personalized e-commerce shopping experience. A shopping trajectory consists of searches, clicks, and purchase records, which contain rich information about customer's preferences and intent. In real-world scenarios, however, these records are rarely accessible due to privacy concerns, which makes it difficult to study them experimentally. Moreover, they often span long time periods, leading to extremely long trajectories whose analysis poses a major challenge for frontier large language models (LLMs) due to their fixed in-context window size. These two obstacles motivate the need for a new benchmark and accompanying training data to study reasoning over long shopping trajectories.

Recent studies have shown that reinforcement learning from human feedback (RLHF) \citep{ouyang2022training} can substantially improve the reasoning abilities of LLMs by encouraging the generation of structured reasoning processes.
Building on this idea, reinforcement learning with verifiable rewards (RLVR) \citep{feng2025retool, xie2025unlocking} 
has emerged as a promising alternative. Instead of relying on a learned reward model, RLVR leverages task-specific, rule-based, verification signals to directly supervise model optimization. By aligning training objectives more closely with downstream tasks, this paradigm has demonstrated strong improvements in domains such as mathematical reasoning and code generation. 

However, despite these advances, effectively applying RLVR in long-context scenarios and in the shopping domain remains challenging. Extending RLVR naively to ultra-long trajectories by simply including the entire sequence into the model’s in-context window is both computationally expensive and inherently limited. As trajectories approach the upper bounds of modern LLMs’ context windows, the number of tokens that must be attended simultaneously grows dramatically, leading to quadratic memory and computation costs. This not only makes training and inference inefficient but also degrades reasoning performance, as the model struggles to retrieve and integrate relevant information from distant parts of the trajectory.

To address this gap, we introduce ShopTrajQA, a long-context resource grounded in realistic shopping scenarios that provides both an evaluation benchmark and accompanying training data. The dataset is constructed using real product catalogs combined with LLM-simulated long-horizon user trajectories. To ensure benchmark reliability, each trajectory is generated with verifiable procedures: the LLM is prompted to produce sequences of realistic actions (e.g., searches, clicks, and purchases) that are consistent with product attributes. For QA tasks, corresponding answers are generated using a strong reference LLM with tool-assisted verification. This pipeline ensures that the resulting trajectories are internally consistent, while enabling each QA example to be objectively validated. ShopTrajQA provides two context-length variants, reaching up to 32k and 64k tokens, enabling systematic evaluation of model robustness under increasing trajectory lengths.

Benchmarking 9 frontier LLMs on ShopTrajQA reveals substantial performance degradation when long shopping trajectories are directly included in the prompts. As input length increases, models struggle to effectively retrieve relevant information and maintain coherent reasoning. To address this challenge, we propose a {\em Customer Agent Framework} that enables models to handle ultra-long contexts through explicit tool interactions. Instead of including the entire trajectory into the prompt, our framework stores context as external local files and allows the agent to selectively retrieve and parse relevant information through tool usage. 
This design improves reasoning in two ways. 
$i)$ Instead of requiring the model to process an entire long trajectory at once, our framework retrieves only small, relevant fragments for reasoning. When a long sequence is directly included in the context window, the model must simultaneously identify relevant information and perform reasoning, which can degrade its ability to effectively retrieve and utilize useful signals. By leveraging tools to selectively retrieve relevant pieces of context, the model can focus more on reasoning rather than searching through large amounts of redundant text. 
$ii)$ By storing trajectories as external files and retrieving them on demand, the framework is no longer bound by the model’s native context-length limit. In principle, it can scale to extremely long contexts, limited only by the capacity of the external storage system rather than by the model context window.

\begin{figure*}[t]
    \centering
    \includegraphics[width=1.7\columnwidth]{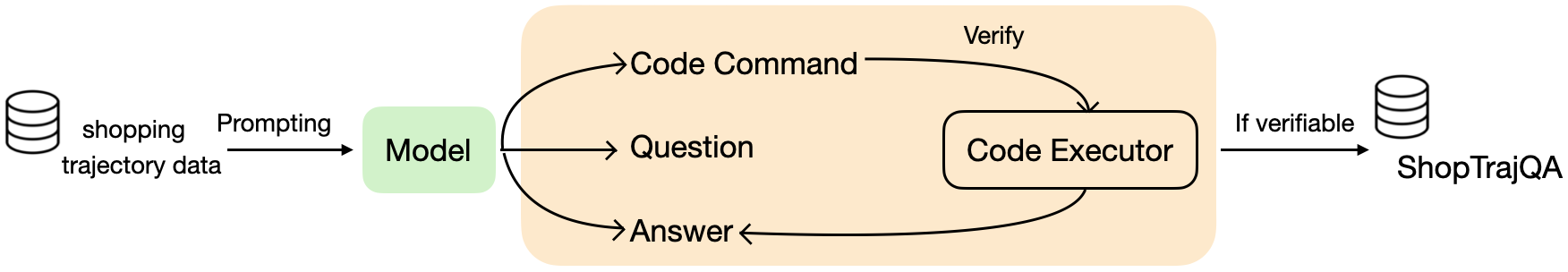}
    \caption{
    Retrieval QA Construction: the model generates and verifies executable QA from shopping trajectory data.
    }
    \vspace{-4mm}
    \label{fig:data_construction}
\end{figure*}

Our contributions are summarized as follows:
%
\begin{itemize}
\item We introduce ShopTrajQA, a novel long-context QA resource for the shopping domain that includes both an evaluation benchmark with 32k and 64k token variants and accompanying training data, the latter comprising a verified supervised fine-tuning (SFT) set of multi-turn tool-use trajectories and a Reinforcement Learning with Verifiable Rewards (RLVR) set with executable, automatically-checkable answers. Extensive benchmarking of mainstream LLMs across different model scales reveals key performance bottlenecks in processing long shopping trajectories.
\item We propose a {\em Customer Agent Framework} that stores trajectories as externalized context files and trains the agent, via an SFT cold start followed by RLVR finetuning, to retrieve and reason over them through code-interpreter interactions. This design effectively overcomes context-length limitations and substantially improves reasoning performance over long documents.
\item Experiments on ShopTrajQA and five additional reasoning benchmarks demonstrate that our framework achieves strong performance and generalization.

\end{itemize}

\section{Related Work}

Reasoning has become increasingly important for improving the capabilities of large language models (LLMs), particularly on complex and knowledge-intensive tasks \citep{wang2025think, qiu2026anchorabductivenetworkconstruction, ji2026strideedstrategygroundedstepwisereasoning, luo2026gcotdecodingunlockingdeepreasoning}. 
Tool-integrated reasoning further provides an effective paradigm for enhancing LLMs by enabling access to external knowledge sources and computational tools. Early approaches such as Retrieval-Augmented Generation (RAG) \citep{lewis2020retrieval, guu2020realm} retrieve relevant documents from external knowledge bases to improve factual grounding, while later retrieval-based frameworks have shown strong performance on knowledge-intensive tasks \citep{luo-etal-2025-dtcrs, xu2026self}. 
However, these methods typically rely on retrieving relevant context before inference, providing the model with information to reason over in a single pass. Recent work shows that allowing models to interact with tools iteratively during reasoning can improve problem solving, as multi-step tool interactions enable models to dynamically retrieve missing information and refine intermediate steps \citep{jin2025search, chen2026learning, qiu2026dimmemdimensionalstructuringefficient, li2026arise}. Reinforcement Learning with Verifiable Rewards (RLVR) has further been used to train models to perform such multi-turn tool interactions.

Beyond retrieval-based question answering, tool-integrated reasoning has also been applied to mathematically and computationally demanding tasks by incorporating programmatic reasoning and external execution \citep{chen2022program, wang2024empowering, zhang2026logicalphasetransitionsunderstanding, zhang2026coupling, zhang2026semanticawarelogicalreasoningsemiotic}. 
Subsequent work studies when tools should be invoked and how invocation strategies affect performance \citep{feng2025retool, li2026reason}. However, learning effective tool use remains challenging due to execution errors and malformed tool calls. \citet{li2025cort} shows that structured tool invocation can improve both tool-use efficiency and reasoning reliability.

\citet{wang2025loongrl} explore the use of reinforcement learning to enhance long-context reasoning by constructing ultra-long training data that encourages models to reason over extended textual inputs. While this approach improves the model's ability to process longer sequences, it relies on the model's native context window, and therefore remains fundamentally constrained by the fixed context-length limitation of the underlying LLM. Reinforcement learning has also been explored in shopping-related tasks. For example, Shop-R1 \citep{zhang2025shop} focuses on modeling shopping behaviors through reinforcement learning, such as simulating online shopping decision-making. Customer-R1 \citep{wang2025customer} further explores incorporating customer personas to simulate shopping behavior.
However, these studies primarily focus on behavior simulation rather than reasoning over long-term interaction histories. In particular, RLVR-based tool-interactive reasoning for extracting answers from long and structured shopping trajectories remains largely unexplored.
Evaluating the outputs of LLMs is critical for their effective use \citep{lan2025f2bench, lan2025mcbe, li2026wrote, wang2026creativebench}. OPeRA \citep{wang2025opera} introduces a public shopping-trajectory benchmark, but its trajectories are relatively short and mainly centered around student users. 

To address this gap, we introduce ShopTrajQA, a long-context resource for shopping-trajectory reasoning, pairing an evaluation benchmark with accompanying SFT and RLVR training data, all constructed from real-world product information and simulated shopping trajectories. We further propose a {\em Customer Agent Framework} that interacts with externalized local files to overcome context-length limitations and enable multi-step reasoning over complex shopping histories.

\section{The ShopTrajQA Dataset}\label{sec:dataset}
In this section, we describe the construction of ShopTrajQA, which we build in three stages that together yield both an evaluation benchmark and its accompanying training data. We first simulate long-horizon shopping trajectories grounded in real-world product information, where each trajectory is a sequence of realistic user actions (searches, clicks, and purchases) consistent with product attributes and plausible behavior patterns (Section~\ref{sec:traj_simulation}). On top of these trajectories, we generate retrieval-based question–answer pairs, each paired with an executable code command (e.g., a SQL query) whose syntax and returned answer are automatically verified by a code interpreter, forming the evaluation benchmark (Section~\ref{sec:QA_Construction}). Finally, from the same verified QA pairs, we derive two training sets: a supervised fine-tuning (SFT) set of multi-turn tool-use trajectories for cold start (Section~\ref{sec:sft_data}), and a Reinforcement Learning with Verifiable Rewards (RLVR) set whose verified answers serve as reward signals (Section~\ref{sec:rlvr_data}). Compared with existing shopping datasets, ShopTrajQA features substantially longer trajectories and greater product diversity, and provides multiple context-length variants that enable systematic evaluation under varying context sizes.

\subsection{Long Shopping Trajectory Simulation} \label{sec:traj_simulation}
OPeRA \citep{wang2025opera} is a public shopping-trajectory dataset that serves as a baseline for this work. However, it has two key limitations. Since the data was collected in an academic setting, the user group is mostly limited to students and researchers. 
As a result, the users in OPeRA have similar income levels and shopping habits, which leads to limited diversity in product categories and shopping behavioral patterns. Furthermore, the trajectories in OPeRA are relatively short, limiting its suitability for evaluating long-context reasoning and behavioral understanding.
Nevertheless, while OPeRA's user demographics are limited, the distribution of user actions (e.g., search, click, and purchase patterns) still provides a useful reference for realistic shopping behaviors. 


To address the above limitations, we construct a new dataset by leveraging product query data from the public ESCI dataset \citep{reddy2022shopping}. ESCI contains around 130 thousand unique queries and 2.6 million manually labeled {\em (query, product)} relevance judgments, together with rich product metadata such as product IDs, titles, and brand information. The large scale and diversity of product information significantly expand the space of possible interactions during trajectory generation. Guided by the action distribution observed in OPeRA, we prompt an LLM to generate long-horizon shopping trajectories using ESCI product query data. The generation follows a predefined schema shown in Appendix~\ref{appendix_shopping_prompt}. The rich product space enables the simulation process to explore a broader range of shopping intents and product categories, leading to more diverse user profiles and trajectories.

To facilitate the evaluation of long-context capabilities, we construct two versions of the dataset with maximum context lengths of 32k and 64k tokens. The 32k version is obtained by truncating trajectories from the 64k version. Figure~\ref{fig:shop_data_construction} illustrates the differences between ShopTrajQA and OPeRA. Compared to OPeRA, ShopTrajQA provides substantially longer trajectories and a more diverse set of product information, making it better suited for studying long-context understanding in the shopping domain. We report the data distribution details in Appendix~\ref{app:benchmark_detail}. 

\begin{figure*}[th]
    \centering
    \includegraphics[width=1.9\columnwidth]{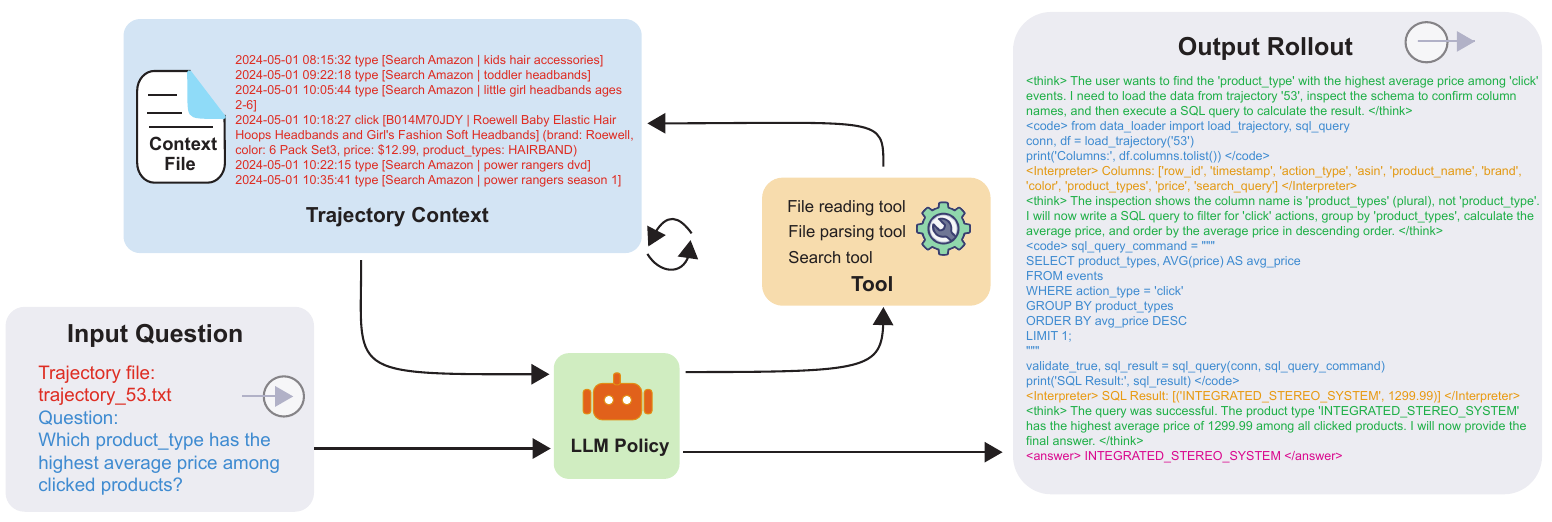}
    \caption{Customer-Agent framework. Red text denotes shopping trajectory information.}
    \vspace{-2mm}
    \label{fig:framework}
\end{figure*}

\subsection{Retrieval QA Construction} \label{sec:QA_Construction}
Based on the generated shopping trajectories, we leverage an LLM to generate verifiable question-answer pairs, which are then used to construct the ShopTrajQA benchmark. Verifiable means that each generated question includes a corresponding executable command ({\em e.g.}, SQL, regex, or other programmatic queries) that can retrieve the answer directly from the underlying shopping trajectory.

The workflow is illustrated in Figure~\ref{fig:data_construction}. We prompt the model to produce a question along with a code command that can be executed to obtain the answer. We then validate each example by running the generated command through a code executor to ensure that $i$) the command executes successfully and $ii$) the returned result matches the provided answer.
By requiring each example to include an executable, verifiable command and validating its correctness, we improve the quality and reliability of the constructed data.
These question-answer pairs constitute a benchmark for evaluating a model’s ability to answer questions in the shopping domain. 
Details of this benchmark are reported in Appendix~\ref{app:benchmark_detail}.

\subsection{SFT Training Data}\label{sec:sft_data}
Beyond the evaluation benchmark, we release a multi-turn supervised fine-tuning (SFT) dataset that bootstraps an agent's tool-use ability as a cold start (its role in training is described in Section~\ref{sec:sft_training}). Building on the retrieval QA pairs above, we prompt an LLM to answer each question through \emph{multi-turn} interactions with a code interpreter, and we retain only those rollouts that ($i$) successfully execute all tool calls and ($ii$) recover the correct final answer from the tool outputs. This yields \textbf{762} verified multi-turn, tool-use trajectories, each pairing a shopping-domain question with an executable reasoning process (tool call, tool response, and final answer). Because every trajectory is grounded in a verifiable execution, the resulting supervision is faithful rather than hallucinated.

\subsection{RLVR Training Data}\label{sec:rlvr_data}
We further release a Reinforcement Learning with Verifiable Rewards (RLVR) dataset of 10{,}822 samples (its use in RL training is detailed in Section~\ref{sec:rlvr}). Each RLVR sample has the same structure as an evaluation-benchmark example: a shopping-domain question and its underlying trajectory stored as an external file. In addition, we attach a reference executable command (e.g., a SQL query) that retrieves the answer and the verified ground-truth answer, which serves as the target reward signal. The key difference from the SFT data is that RLVR samples do not include a demonstration trajectory. Rather than imitating a reference reasoning process, the agent is trained to autonomously generate the command that retrieves the answer from the external file, and is optimized directly against the correctness of its final answer. Together, the SFT and RLVR sets form a complete training suite for building agents capable of solving ShopTrajQA.

Table~\ref{tab:dataset_summary} summarizes the released artifacts. ShopTrajQA comprises two training datasets (SFT and RLVR) and two long-context evaluation sets (32k and 64k tokens). All data are constructed from public product information and synthetic trajectories, and are released publicly to support reproducibility.

\begin{table}[t]
\centering
\caption{Overview of the released ShopTrajQA artifacts. All subsets are in the shopping domain. Evaluation sets differ in trajectory context length.}
\label{tab:dataset_summary}
\setlength{\tabcolsep}{5pt}
\scalebox{0.9}{\begin{tabular}{llrl}
\toprule
\textbf{Dataset} & \textbf{Split} & \textbf{\#Samples} & \textbf{Purpose} \\
\midrule
\rowcolor{gray!12}
\multicolumn{4}{l}{\textit{Training data}} \\
ShopTrajQA-SFT   & train & 762    & Cold-start tool-use SFT \\
ShopTrajQA-RLVR  & train & 10{,}822 & RL with verifiable rewards \\
\midrule
\rowcolor{gray!12}
\multicolumn{4}{l}{\textit{Evaluation data}} \\
ShopTrajQA-32k   & test  & 100    & Long-context eval (32k) \\
ShopTrajQA-64k   & test  & 100    & Long-context eval (64k) \\
\bottomrule
\end{tabular}}
\end{table}

\section{Method}
In this section, we present our agentic framework designed to handle long contexts. We begin by formalizing the inherent limitations of current models when processing long contexts. To address these constraints, we propose a mechanism that stores context as local files and trains the model to interact with these external resources via tool-use, thereby enhancing both efficiency and reliability. We then describe how we use the SFT dataset (Section~\ref{sec:sft_data}) to provide a cold start for tool calling, and finally elaborate on the details of RLVR finetuning with the RLVR dataset (Section~\ref{sec:rlvr_data}). Figure~\ref{fig:framework} provides an overview of the agent framework.

\subsection{Problem Definition}\label{sec:problem_definition}
Let $G = \pi_\theta(P)$ denote the process of prompting a policy model $\pi_\theta$ to generate a response $G$. Typically, a prompt $P$ consists of two components: the \textit{context} ({\em e.g.}, a customer shopping trajectory) and a \textit{question}. 
However, as the \textit{context} length scales, inference efficiency degrades substantially and may eventually surpass the architectural constraints of the model's maximum context window.
To address this issue, we decouple the context from the question by offloading the former to local storage. In this paradigm, the model functions as an agent that interacts with local context files through code-interpreter–integrated reasoning. By programmatically accessing externalized files, the model selectively retrieves relevant information to support its reasoning process, thereby effectively bypassing fixed context-length constraints. Consequently, it is essential for the model to acquire this integrated reasoning and tool-use capability. 

We formulate code-interpreter-integrated reasoning as a reinforcement learning problem with interleaved tool execution. 
The system consists of a policy model $\pi_\theta$, which generates reasoning text and code snippets, 
a code interpreter $\mathrm{CI}$, which executes the generated code and returns results or errors, 
and an environment that provides the question $q$ along with ground-truth answers $a$.

At each time step $t$, the policy model generates either the reasoning text $g_t$ or the code $c_t$.
When the code termination trigger \texttt{</code>} is detected, generation pauses, the code within \texttt{<code> </code>} tags
is sent to the code interpreter $\mathrm{CI}$, and feedback $f_t$ (results or errors) is
returned within the \texttt{<interpreter></interpreter>} tags.
The feedback $f_t$ is concatenated with the most recent generation $g_t \oplus c_t$ 
and fed back to the model as input to generate the next reasoning text $g_{t+1}$. 
This process continues until the model produces the final answer $o$ without detecting code, forming a hybrid trajectory:
\[
\tau = \left[g_1 \oplus c_1 \oplus f_1 \oplus g_2 \oplus c_2 \oplus f_2 \oplus \cdots \oplus o \right].
\]

At each time step $t$, the model receives a reward $r_t$. 
The objective is to maximize the expected cumulative reward over the trajectory. 
Let $T$ denote the total number of time steps in an interaction trajectory. 
The first $T-1$ steps correspond to tool-invocation turns, while the final step evaluates the correctness of the generated answer.


Rewards are assigned based on two aspects: the executability of intermediate tool calls  at each time step $t$ and the correctness of the final answer $o$. 
Through reinforcement learning, we enable the model to autonomously leverage tools to access and parse local context files, thereby developing the ability to perform complex long-context understanding. The following sections detail how this paradigm is instantiated: we first describe the architectural components that realize tool-integrated reasoning, then outline the two-stage optimization process consisting of SFT cold-start initialization followed by RLVR fine-tuning. 

\subsection{Customer Agent Framework}\label{sec:customer_agent}
\noindent{\bf Local Storage Mechanism} Placing long texts directly in the prompt leads to a noticeable degradation in a model’s memory and reasoning ability \citep{du2025context, liu2024lost, kuratov2024babilong}. Simply making the input excessively long can cause the model to fail to retain or effectively use all the information. Moreover, when the model is required to perform tool-based operations such as retrieval or parsing on long texts, it often results in missing relevant information. In multi-turn settings, the time and computational cost introduced by long inputs can further grow multiplicatively. To address this issue, we integrate the data loading and parsing tool into the sandbox environment, as illustrated in Figure~\ref{fig:framework}. This tool allows the model to directly access datasets by their IDs, automatically parse the data according to predefined formats, and construct a database. In addition, it enables the model to retrieve key information using simple SQL queries.

\noindent{\bf Interpreter Feedback Masking}
Tokens within \texttt{<interpreter> </interpreter>} tags are not generated by the policy and are therefore masked out from loss computation. This prevents external feedback tokens from degrading the coherence of model-generated reasoning.

\noindent{\bf KV-Cache Reuse}
When code execution is triggered, we cache key-value (KV) states before execution and compute new KV states only for the interpreter feedback, reducing memory overhead.

\noindent{\bf Asynchronous Sandbox}
We use sandboxed code execution to support safe and scalable multi-turn execution-driven training.
The sandbox isolates model-generated code through strict system-access and resource constraints, which is especially important in reinforcement learning settings with exploratory code generation. By decoupling code execution from model inference, the sandbox enables parallel execution without blocking GPU computation, substantially improving throughput and scalability for large-scale training and evaluation.

\subsection{SFT Cold-Start Training}\label{sec:sft_training}
Directly applying RL to models with no prior tool-use experience often results in prohibitive exploration costs, as the model struggles to generate well-formatted code snippets. Such frequent syntax errors can trap the agent in suboptimal cycles or execution deadlocks \citep{feng2025retool}. Consequently, a more efficient strategy is to initiate a cold start via SFT training. This stage equips the model with a foundational understanding of tool-calling syntax, thereby establishing a robust starting policy for subsequent RLVR optimization.

We use the SFT dataset introduced in Section~\ref{sec:sft_data}, which consists of verified multi-turn, tool-use rollouts derived from the retrieval QA pairs. The key idea is that by training on real, verifiable multi-turn tool-use rollouts, the model learns how to correctly invoke tools, interpret tool responses, and derive the final answer based on the returned response, establishing a robust starting policy for subsequent RLVR optimization.

\subsection{Reinforcement Learning}\label{sec:rlvr}
Drawing inspiration from the success of DeepSeek-R1 \citep{guo2025deepseek}, we adopt \emph{Group Relative Policy Optimization} (GRPO).
For each question $q$ drawn from the training question distribution $P(Q)$, GRPO samples a group of $G$ outputs
$\{o_1, o_2, \ldots, o_G\}$ from the old policy $\pi_{\theta_{\mathrm{old}}}$ and optimizes
the policy model $\pi_\theta$ by maximizing:

\begin{equation}
\scalebox{0.9}{$
\begin{aligned}
\mathcal{J}_{\mathrm{GRPO}}(\theta)
= \;&
\mathbb{E}_{q \sim P(Q),\, \{o_i\}_{i=1}^{G} \sim \pi_{\theta_{\mathrm{old}}}(\cdot \mid q)}
\Bigg[
\frac{1}{G}
\sum_{i=1}^{G}
\Big(
\\
&\min\!\Big(
r_i(\theta) A_i,\;
\operatorname{clip}\!\big(
r_i(\theta), 1-\epsilon, 1+\epsilon
\big) A_i
\Big)
\\
&\quad
- \beta\, D_{\mathrm{KL}}\!\left(
\pi_{\theta}(\cdot|q) \,\|\, \pi_{\mathrm{ref}}(\cdot|q)
\right)
\Big)
\Bigg]
\end{aligned}
$}
\end{equation}

where
\[
r_i(\theta)=
\frac{\pi_{\theta}(o_i \mid q)}
{\pi_{\theta_{\mathrm{old}}}(o_i \mid q)}
\]
is the importance sampling ratio between the current policy and the behavior policy. The normalized advantage is computed as
\[
A_i=
\frac{R_i-\operatorname{mean}(\{R_j\}_{j=1}^{G})}
{\operatorname{std}(\{R_j\}_{j=1}^{G})},
\]
where $R_i$ denotes the reward of the $i$-th rollout, and the normalization is performed within each rollout group. Here, $G$ is the number of sampled rollouts, $\epsilon$ is the clipping coefficient, and $\beta$ controls the strength of the Kullback--Leibler (KL) regularization term
$D_{\mathrm{KL}}(\pi_{\theta}(\cdot\mid q)\,\|\,\pi_{\mathrm{ref}}(\cdot\mid q))$.
The KL loss is added to prevent the policy from deviating excessively from the reference policy $\pi_{\mathrm{ref}}$.




\noindent{\bf Rewards}
As described in Section~\ref{sec:problem_definition}, the reward consists of two components: $i$) the executability of intermediate code at first $T-1$ step, and $ii$) the correctness of the final answer $o$. 
Specifically, to stabilize training and improve convergence, we adopt the reward function proposed in \cite{jiang2025verltool}, which combines answer correctness with tool usage.
The correctness component evaluates the predicted answer against the ground truth in a strict format (e.g., within \texttt{\textbackslash boxed\{\}}). 
To encourage useful exploration, we provide an additional reward proportional to the number of executable tool calls.
Formally, the reward is defined as
\[\scalebox{1}{$
R(a, \hat{a}, T) =
\begin{cases}
1, & \text{if } \hat{a} = a \\
\min(\alpha, -1 + 0.1 \cdot (T-1)), & \text{otherwise}
\end{cases}$}
\]
where $a$ and $\hat{a}$ denote the ground-truth and predicted answers in the strict format  extracted from the final output $o$, respectively, and $T-1$ denotes the number of executable tool-invocation turns.
If the model fails to produce the correct answer, the reward reflects the quality of tool usage. 
When no executable tool calls are generated, the model receives a reward of $-1$. 
Additional executable tool calls yield incremental rewards, encouraging meaningful tool interaction. 
However, to prevent the number of tool calls from dominating the reward signal, the reward is capped at a hyper parameter $\alpha$. 
To avoid additional computational overhead, we use exact match to determine correctness.
This designed reward reduces reward hacking, promotes diverse problem-solving strategies, and encourages the model to explore tool usage even when the final answer is incorrect.

\section{Experiments}\label{section:experiment}

\noindent{\bf Datasets}
We evaluate our method on six datasets spanning mathematical reasoning, shopping behavioral reasoning, and multi-hop QA tasks:

\noindent\textit{Mathematical Reasoning:} We use AIME 2024 and AIME 2025, challenging mathematical competition benchmarks requiring multi-step reasoning.
\noindent\textit{Shopping Behavioral Reasoning:} We evaluate on ShopTrajQA, our long-context dataset built from customer shopping trajectories. ShopTrajQA contains verifiable QA pairs in which answers can be retrieved from shopping trajectories using executable code commands. We include two context-length variants: a short-context version with 32k tokens and a long-context version with 64k tokens.
\noindent\textit{Multi-hop QA:} To assess generalization to complex reasoning tasks, we evaluate on three multi-hop question answering benchmarks: 2WikiMultihopQA~\citep{ho2020constructing}, HotpotQA~\citep{yang2018hotpotqa}, and Bamboogle~\citep{press2022bamboogle}. These datasets require models to integrate information from multiple sources to answer questions.
\noindent\textit{Training Data:} It consists of the SFT data described in Section~\ref{sec:sft_training} and the RL data without tool-use traces. Appendix~\ref{app:dataset_summary} reports details of these datasets.

\begin{figure}[t]
\centering
\includegraphics[width=1\columnwidth]{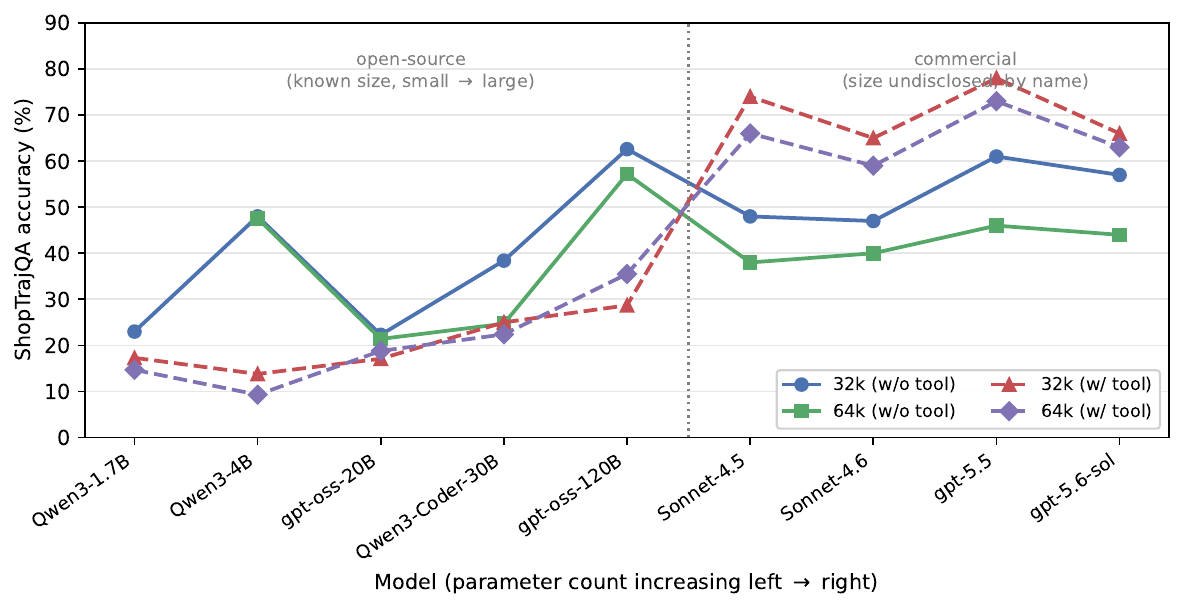}
\caption{Benchmark accuracy performance on ShopTrajQA evaluation set versus model scale (untrained models). Models are ordered by increasing parameter count ($x$-axis); commercial models, whose sizes are undisclosed, are ordered by name to the right of the divider. Each line is one inference setting (32k/64k $\times$ with/without tool).} 
\label{fig:param_vs_perf}
\end{figure}

\noindent{\bf Baselines}
We apply our method to Qwen models of different sizes, including \texttt{Qwen3-1.7B} and \texttt{Qwen3-4B}~\citep{yang2025qwen3}.
To examine whether larger models exhibit stronger tool-calling and question-answering capabilities, we also compare against three popular large-scale LLMs: \texttt{gpt-oss-20B}, \texttt{Qwen3-Coder-30B}, and \texttt{gpt-oss-120B}~\citep{yang2025qwen3, openai2025gptoss}.

\vspace{1mm}
\noindent{\bf Evaluation Metrics}\label{sec:eval_metric}
We evaluate the performance of the model using the following metrics.

\noindent\textit{Accuracy:} We quantify accuracy using pass@1, which reflects whether the model produces a correct answer in its top-ranked generation. For AIME dataset, pass@1 is computed over 8 independent repetitions to reduce variance, and Exact Match is used to determine answer correctness. For ShopTrajQA, considering the large scale of the dataset, we conduct three evaluation run for efficiency. Due to the diversity of possible answers, we use an LLM-based judge to determine correctness; the evaluation prompt is provided in Appendix~\ref{app:prompt_llmjudge}.

\noindent\textit{Multi-hop QA Metrics:} For 2WikiMultihopQA, HotpotQA, and Bamboogle, we report both exact match (EM) and F1 scores, which are standard metrics for evaluating question answering systems.

\noindent\textit{Tool Usage Metrics:} To measure tool-calling capability, we define Tool Call Rate (TCR) as the proportion of instances where the model attempts to invoke at least one external tool during reasoning. The successful tool call rate (SuccTCR) measures the proportion of instances in which the model successfully executes at least one tool call. In our setting, essential contextual information must be retrieved via external tools. Therefore, both TCR and SuccTCR are monotonically beneficial: higher values indicate stronger awareness of tool necessity and more reliable tool execution, leading to more complete context acquisition and improved downstream reasoning.

\noindent{\bf Implementation}
Our method is trained on eight NVIDIA H100 GPUs using \texttt{Qwen3-1.7B} and \texttt{Qwen3-4B} as the base models.
Following the procedure described in Section~\ref{sec:sft_training}, we collect an SFT dataset consisting of tool-use reasoning trajectories.
We first apply SFT on this dataset as a cold start, enabling the model to learn when and how to invoke a code interpreter for computation.
We then perform RLVR training to further refine this capability.
In the following sections, we report results for both the SFT and RLVR versions of the models.
The hyperparameter selection and their settings are described in Appendix~\ref{appendix:hyperparams}.

\begin{table}[t]
\centering
\caption{Training performance of our models on ShopTrajQA under different context lengths. Results show accuracy (\%). ``--'' indicates that the prompt token length exceeds the maximum context window.}
\scalebox{1.1}{
\begin{tabular}{lcc}
\toprule
Model & 32k & 64k \\
\midrule
\rowcolor{gray!12}
\multicolumn{3}{c}{\textbf{\textit{Qwen3-1.7B}}} \\
Qwen3-1.7B (w/o tool) & 23.0 $\pm$ 0.6 & -- \\
Qwen3-1.7B (w/ tool) & 17.3 $\pm$ 0.5 & 14.7 $\pm$ 0.6 \\
Qwen3-1.7B-SFT & 28.1 $\pm$ 0.7 & 27.7 $\pm$ 0.8 \\
Qwen3-1.7B-SFT-RLVR & \textbf{59.2 $\pm$ 1.1} & \textbf{51.0 $\pm$ 1.0} \\
\midrule
\rowcolor{gray!12}
\multicolumn{3}{c}{\textbf{\textit{Qwen3-4B}}} \\
Qwen3-4B (w/o tool) & 48.0 $\pm$ 1.0 & 47.6 $\pm$ 1.1 \\
Qwen3-4B (w/ tool) & 13.8 $\pm$ 0.4 & 9.3 $\pm$ 0.5 \\
Qwen3-4B-SFT & 36.4 $\pm$ 0.9 & 41.0 $\pm$ 1.0 \\
Qwen3-4B-SFT-RLVR & \textbf{62.5 $\pm$ 1.3} & \textbf{60.1 $\pm$ 1.2} \\
\bottomrule
\end{tabular}}
\vspace{-2mm}
\label{tab:training_results}
\end{table}

\begin{table*}[t]
\centering
\caption{Model performance comparison across AIME mathematics and Multi-hop QA benchmarks.} 
\scalebox{1}{
\setlength{\tabcolsep}{4pt}
\begin{tabular}{l|cc|cc|cc|cc}
\toprule
\multirow{2}{*}{Model} & \multicolumn{2}{c|}{AIME  (\%)} & \multicolumn{2}{c|}{2wiki} & \multicolumn{2}{c|}{HotpotQA} & \multicolumn{2}{c}{Bamboogle} \\
 & AIME24 & AIME25 & EM & F1 & EM & F1 & EM & F1 \\
\midrule
Qwen3-1.7B & 32.08 & \textbf{28.33} & 0.156 & 0.200 & 0.142 & 0.206 & 0.108 & 0.102 \\
Qwen3-1.7B-SFT-RLVR & 32.08 & 27.50 & 0.156 & \textbf{0.201} & 0.142 & \textbf{0.207} & \textbf{0.112} & \textbf{0.170} \\
\midrule
Qwen3-4B & 62.08 & 47.50 & \textbf{0.336} & 0.405 & \textbf{0.307} & \textbf{0.402} & 0.356 & 0.439 \\
Qwen3-4B-SFT-RLVR & \textbf{65.83} & \textbf{55.00} & 0.331 & \textbf{0.407} & 0.306 & 0.400 & \textbf{0.368} & \textbf{0.459} \\
\bottomrule
\end{tabular}}
\label{tab:model_comparison}
\end{table*}

\subsection{Results and Analysis}

\noindent{\bf ShopTrajQA Benchmark Performance}
We first benchmark 9 frontier models without task-specific training on ShopTrajQA evaluation set under the 32k and 64k context settings. Figure~\ref{fig:param_vs_perf} summarizes the results as a function of model scale, covering commercial models (gpt-5.5, gpt-5.6-sol, Sonnet-4.5, and Sonnet-4.6) and open-source models (gpt-oss-120B, Qwen3-Coder-30B, gpt-oss-20B, Qwen3-4B, and Qwen3-1.7B); the full per-model numbers with standard deviations are reported in Appendix~\ref{app:benchmark_table} (Table~\ref{tab:qwen3_results}). Each model supports two inference modes: $i$) pure text inference, where the entire shopping trajectory is included in the prompt, and $ii$) tool-calling inference, where the model reads local shopping trajectory files, parses them and obtains answers via SQL execution.

These results reveal two gaps that motivate our approach.

\noindent\textit{Gap 1: Tool-calling capability.} At a fixed context length, the two model groups respond to tool use in opposite ways. For all four commercial models, tool calling yields their best accuracy and consistently outperforms pure text: gpt-5.5 improves from 61.0 to 78.0 at 32k and from 46.0 to 73.0 at 64k, Sonnet-4.5 improves from 48.0 to 74.0 and from 38.0 to 66.0, and even gpt-5.6-sol, the weakest tool user among them, improves from 57.0 to 66.0 and from 44.0 to 63.0, respectively. This indicates that strong models can readily adapt to our code-interpreter tool and exploit it for retrieval. In contrast, tool calling consistently underperforms pure text across every open-source model; for instance, gpt-oss-120B drops from 62.6 to 28.7 at 32k, and Qwen3-4B drops from 48.0 to 13.8. Since the trajectory is stored as an external local file and its content is only accessible through tool interactions, an open-source model that is not well suited to our tool-use scenario cannot reliably invoke the tool, parse the returned data, or manage execution state across code blocks, and therefore fails to reach the correct answer. The accuracy under tool calling thus reflects not only answer correctness but also the model's ability to use tools, and this gap is precisely what motivates explicitly training smaller models for tool-calling capability. We provide representative case studies in Appendix~\ref{app:case_studies}. 

\noindent\textit{Gap 2: Long-context limitation.} For a fixed model, the 64k setting is consistently more challenging than 32k under both inference modes; for example, gpt-5.5 drops 15.0 points and Qwen3-Coder-30B drops 13.7 points in pure text as the context grows from 32k to 64k. This degradation reflects a long-context bottleneck: as trajectories lengthen, models increasingly struggle to retrieve relevant information and sustain coherent reasoning, motivating a framework that decouples reasoning from the raw context window.

\noindent{\bf Training Performance}
To enable smaller open-source models to learn correct reasoning through tool use, we conduct SFT and RLVR training on Qwen3-1.7B and Qwen3-4B. Table~\ref{tab:training_results} reports the accuracy of the untrained baseline, the SFT cold-start version, and the final SFT-RLVR version for each base model. The results show substantial improvements after RLVR training.

We observe that in the short-context setting (32k tokens), the Qwen3-1.7B model improves from 17.3 to 59.2 after RLVR, corresponding to a gain of 41.9 points, while the Qwen3-4B model improves from 13.8 to 62.5, corresponding to a gain of 48.7 points. Under the long-context setting (64k tokens), the Qwen3-1.7B model gains 36.3 points, and the Qwen3-4B model gains 50.8 points.
These results show that our method achieves substantial improvements even in long-context scenarios, where pure text reasoning becomes more challenging. Tool usage helps alleviate this difficulty by enabling structured access to the trajectory data. Appendix~\ref{app:training_progress} visualizes this stage-wise progression for both base models.

\noindent{\bf Generalization Performance}
To assess whether training on shopping domain data causes overfitting, we evaluate our models on general mathematical reasoning benchmarks (AIME 2024/2025) and multi-hop QA datasets (2WikiMultihopQA, HotpotQA, Bamboogle). Table~\ref{tab:model_comparison} presents the results.

For mathematical reasoning, we observe no substantial degradation on AIME 2024 or AIME 2025 after training on the ShopTrajQA dataset. The Qwen3-1.7B model exhibits only a minor fluctuation of approximately one point. Interestingly, the Qwen3-4B model trained with RLVR achieves accuracy improvements of 3.75 points on AIME 2024 (from 62.08\% to 65.83\%) and 7.5 points on AIME 2025 (from 47.5\% to 55\%) relative to the baseline. These results indicate that our RLVR training not only preserves, but potentially improves, general reasoning capabilities.
For multi-hop QA benchmarks, our RLVR-trained models maintain competitive performance compared to the baseline models. The Qwen3-4B-SFT-RLVR model achieves the best F1 scores on all three multi-hop QA datasets, demonstrating generalization to complex reasoning tasks that require integrating information from multiple sources. Overall, these results demonstrate that our approach does not suffer from domain overfitting and successfully generalizes across diverse reasoning tasks. 



\noindent{\bf Tool Calling Capability}
To assess the model's tool-calling capability, we evaluate the tool call rate (TCR) and successful tool call rate (SuccTCR), as described in Section~\ref{sec:eval_metric}. We expect both metrics to approach 100\% for optimal performance. Table~\ref{tab:shoppingqa_toolcall} presents tool-calling statistics on ShopTrajQA for both 32k and 64k contexts.

\begin{table}[t]
\centering
\caption{Tool-calling statistics on ShopTrajQA under 32k and 64k contexts. TCR denotes the proportion of instances where a tool call is attempted; SuccTCR denotes the proportion where tool calls execute successfully.}
\scalebox{0.9}{
\begin{tabular}{lcc}
\toprule
Model & TCR & SuccTCR \\
\midrule
\rowcolor{gray!12}
\multicolumn{3}{c}{\textbf{\textit{32k context}}} \\
Qwen3-1.7B (w/o tool)        & 0.00   & 0.00   \\
Qwen3-1.7B (w/ tool)       & 0.76   & 0.42   \\
Qwen3-1.7B-SFT               & 0.68   & 0.57   \\
Qwen3-1.7B-SFT-RLVR          & \textbf{0.85} & \textbf{0.79} \\
\midrule
Qwen3-4B (w/o tool)          & 0.00   & 0.00   \\
Qwen3-4B (w/ tool)         & 0.86   & \textbf{0.79}   \\
Qwen3-4B-SFT                 & 0.79   & 0.77   \\
Qwen3-4B-SFT-RLVR            & \textbf{0.95} & 0.69 \\
\midrule
\rowcolor{gray!12}
\multicolumn{3}{c}{\textbf{\textit{64k context}}} \\
Qwen3-1.7B (w/o tool)        & 0.00   & 0.00   \\
Qwen3-1.7B (w/ tool)       & 0.74   & 0.44   \\
Qwen3-1.7B-SFT               & 0.69   & 0.61   \\
Qwen3-1.7B-SFT-RLVR          & \textbf{0.77} & \textbf{0.73} \\
\midrule
Qwen3-4B (w/o tool)          & 0.00   & 0.00   \\
Qwen3-4B (w/ tool)         & 0.77   & 0.73   \\
Qwen3-4B-SFT                 & 0.84   & \textbf{0.81}   \\
Qwen3-4B-SFT-RLVR            & \textbf{0.93} & 0.66 \\
\bottomrule
\end{tabular}}
\label{tab:shoppingqa_toolcall}
\vspace{-2mm}
\end{table}

For both context lengths, our RLVR-trained models achieve a higher TCR compared to baseline models, indicating greater awareness of when tool use is necessary. The Qwen3-4B-SFT-RLVR model achieves TCR of 0.95 and 0.93 for 32k and 64k contexts, respectively, demonstrating that the model learns to invoke tools in nearly all instances. SuccTCR also improves substantially after RLVR training, confirming that the model learns not only when to call the tools but also how to invoke them correctly. The Qwen3-4B trained with RLVR shows a slightly lower SuccTCR compared to its SFT counterpart. This is likely because RL training encourages more frequent tool calls, leading to a larger number of attempts and consequently a higher probability of execution errors.

\noindent{\bf Data quality evaluation}
Additional analyses on data quality and training-data effectiveness are provided in Appendix~\ref{appendix:data_quality}. 
We sample trajectories from OPeRA and ShopTrajQA to form pairwise comparisons and ask \texttt{GPT-4.1} to judge which is more likely generated from real human behavior; \texttt{GPT-4.1} prefers ShopTrajQA in 73 out of 100 cases. 
We also compare RLVR training using ShopTrajQA against other tool-use datasets, including the SFT and RL data from DAPO-Math-17k \citep{yu2025dapo}; models trained on ShopTrajQA achieve consistently better performance on ShopTrajQA tasks.

\vspace{-1mm}

\section{Discussion}\label{sec:discussion}


In this work, we introduce ShopTrajQA, a long-context resource for reasoning over realistic shopping trajectories that pairs an evaluation benchmark with SFT and RLVR training data. Through extensive experiments, we show that existing LLMs struggle to effectively process long behavioral histories when the entire trajectory is directly included in the prompt.
To address this challenge, we propose a Customer Agent Framework that externalizes long trajectories as structured files and enables models to interact with them through tool usage. Combined with RLVR training on verifiable tool-use trajectories, the framework allows models to retrieve, parse, and reason over long behavioral sequences more effectively.
Experimental results demonstrate that this approach substantially improves reasoning accuracy, particularly under long-context settings. These findings suggest that integrating tool-assisted reasoning is a promising direction for enabling LLMs to operate on complex, long-horizon interaction data. 
\newpage

\bibliographystyle{ACM-Reference-Format}
\bibliography{sample-base}

\appendix

\section{Ethics and Privacy Statement}
Our work does not involve any real user data, personally identifiable information (PII), or human subjects. ShopTrajQA, including both its evaluation and training subsets, is constructed entirely from publicly available resources and synthetic simulation. Specifically, product information is drawn from the public ESCI dataset~\citep{reddy2022shopping}, which contains anonymized product queries and metadata (product IDs, titles, and brand information) and no user identities. The shopping trajectories are not collected from real customers; instead, they are simulated by an LLM, guided only by the aggregate, population-level action distribution (e.g., relative frequencies of searches, clicks, and purchases) reported in the public OPeRA dataset~\citep{wang2025opera}. No individual user record, session, or behavioral log from OPeRA or any e-commerce platform is used, reproduced, or re-identified. Consequently, the dataset raises no concerns regarding informed consent or the exposure of private customer data.

Because the trajectories are synthetic, we explicitly discuss their representativeness rather than claiming fidelity to real behavior. As a limitation, LLM-simulated trajectories may not fully capture the diversity of real-world shopping patterns and can inherit biases from the underlying generation model and from the ESCI product distribution (e.g., skew toward certain product categories or query styles). We mitigate quality risks by requiring every question--answer pair to be accompanied by an executable, verifiable command whose output is automatically validated (Section~\ref{sec:QA_Construction}), but we do not claim that the simulated population is demographically representative of any real customer base. We therefore recommend that ShopTrajQA be used as a controlled benchmark for long-context reasoning rather than as a source of insights about real consumer behavior.

Regarding potential misuse, the released models and data are intended for research on long-context reasoning and tool-augmented agents. As no real user data is involved, the risk of privacy leakage through the released artifacts is minimal. To support transparency and reproducibility, we release the dataset, code, and documentation (including data-collection and preprocessing details) as open-source resources, and we encourage responsible use consistent with the licenses of the underlying public datasets.

\section{Model Hyperparameters}
\label{appendix:hyperparams}

Table~\ref{tab:sft_hyperparams} and Table~\ref{tab:rl_hyperparams} summarize the hyperparameter configurations used for supervised fine-tuning (SFT) and reinforcement learning (RL) training, respectively.

\begin{table}[h]
\centering
\small
\scalebox{1}{
\begin{tabular}{ll}
\toprule
\textbf{Category} & \textbf{Setting} \\
\midrule
Optimizer & AdamW \\
Training epochs & $4$ \\
Training batch size & $32$ \\
Micro-batch size / GPU & $4$ \\
Max sequence length & $16384$ \\
Multi-turn training & Enabled \\
Message field & \texttt{messages} \\
Tool specification field & \texttt{tools} \\
Parallel strategy & FSDP \\
Sequence parallel size & $4$ \\
Padding removal & Enabled \\
Checkpoint save frequency & $20$ steps \\
GPUs & $8$ \\
Nodes & $1$ \\
\bottomrule
\end{tabular}}
\caption{SFT model hyperparameters.}
\label{tab:sft_hyperparams}
\end{table}

\begin{table}[h]
\centering
\small
\scalebox{1}{
\begin{tabular}{ll}
\toprule
\textbf{Category} & \textbf{Setting} \\
\midrule
Advantage estimator & GRPO \\
KL in reward & Disabled \\
KL coefficient & $0.0$ \\
Optimizer learning rate & $1\times10^{-6}$ \\
Training epochs & $15$ \\
Training batch size & $32$ \\
Max prompt length & $2048$ \\
Max response length & $4096$ \\
Max model length & $6144$ \\
Overlong prompt filtering & Enabled \\
PPO mini-batch size & $32$ \\
PPO micro-batch size / GPU & $8$ \\
Clip ratio $\epsilon$ (low, high) & $(0.2, 0.28)$ \\
Clip constant & $10.0$ \\
Loss function $\alpha$ & -0.6 \\
Rollout engine & vLLM (async) \\
Tensor parallel size & $1$ \\
Rollouts per prompt & $8$ \\
Multi-turn rollout & Enabled \\
Max user / assistant turns & $16 / 16$ \\
Validation sampling & Top-$p=0.6$, Temp.=1.0 \\
GPUs & $8$ \\
Nodes & $1$ \\
T maximum & $10$ \\
\bottomrule
\end{tabular}}
\caption{RL Training model hyperparameters.}
\label{tab:rl_hyperparams}
\end{table}




\section{ShopTrajQA Benchmark Details} \label{app:benchmark_detail}

The ShopTrajQA evaluation benchmark is constructed to evaluate models on long-context understanding over shopping trajectories. Each example in the benchmark consists of: (i) a simulated shopping trajectory, (ii) a retrieval-verifiable question, (iii) an executable query (SQL, regex, or other programmatic command), and (iv) the verified answer.

\subsection{Factuality and Verification}

To ensure the reliability of the benchmark, we adopt the following procedures:

\begin{itemize}
\item \textbf{Executable Verification:} Each question includes a programmatic command that retrieves the answer from the underlying trajectory. Commands are automatically executed to confirm they run correctly and return the expected result.
\item \textbf{LLM-Assisted Factuality:} The dataset construction leverages LLMs to generate trajectories and questions. We quantify factual correctness by comparing generated answers against the trajectory.
\item \textbf{Multi-Level Validation:} Commands that fail to execute or produce inconsistent answers are filtered out, ensuring only high-quality, verifiable QA pairs are included.
\end{itemize}

\subsection{Dataset Variants}

ShopTrajQA provides multiple variants to evaluate long-context understanding:

\begin{itemize}
\item \textbf{32k-token Trajectories:} Truncated version for medium-length context evaluation.
\item \textbf{64k-token Trajectories:} Full-length trajectories, challenging models on extreme long-context understanding.
\end{itemize}

This benchmark supports quantifying model performance in terms of retrieval accuracy, factual correctness, and long-context understanding.

\subsection{Data Distribution}\label{appendix:data_distri}

\begin{figure}[th]
    \centering
    \includegraphics[width=\columnwidth]{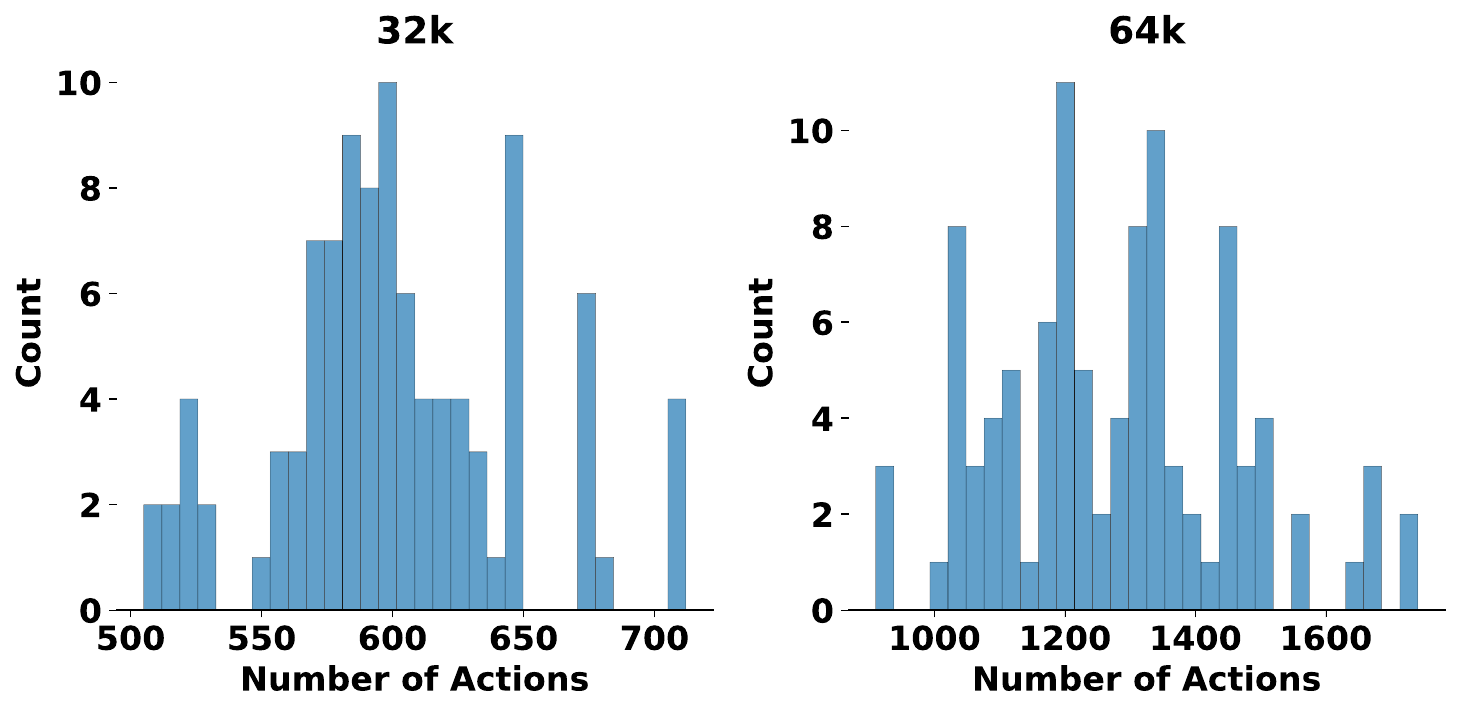}
    \caption{
    Number of action distribution for 32k and 64k setting.
    }
    \label{fig:action_hist}
\end{figure}

\begin{figure}[th]
    \centering
    \includegraphics[width=\columnwidth]{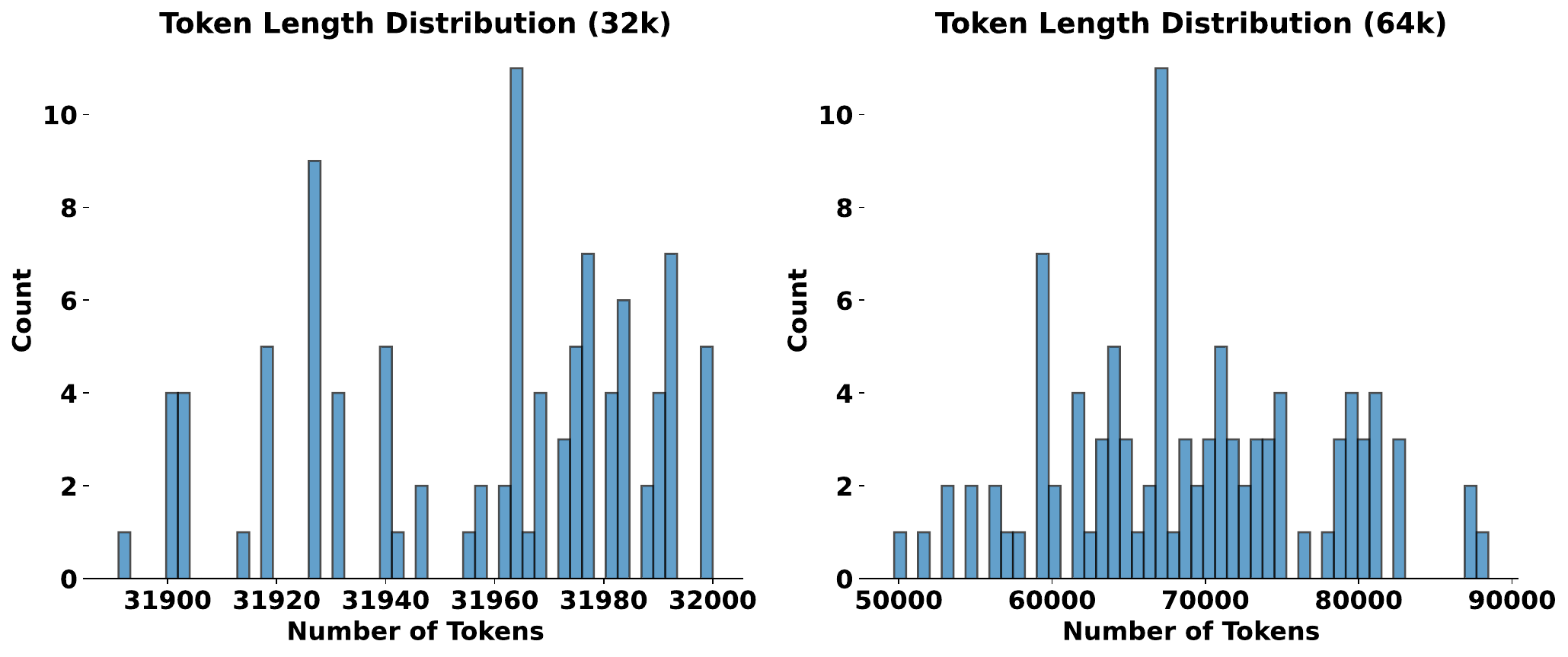}
    \caption{
    Token length distribution plot for 32k and 64k setting.
    }
    \label{fig:token_hist}
\end{figure}

\begin{figure}[th]
    \centering
    \includegraphics[width=\columnwidth]{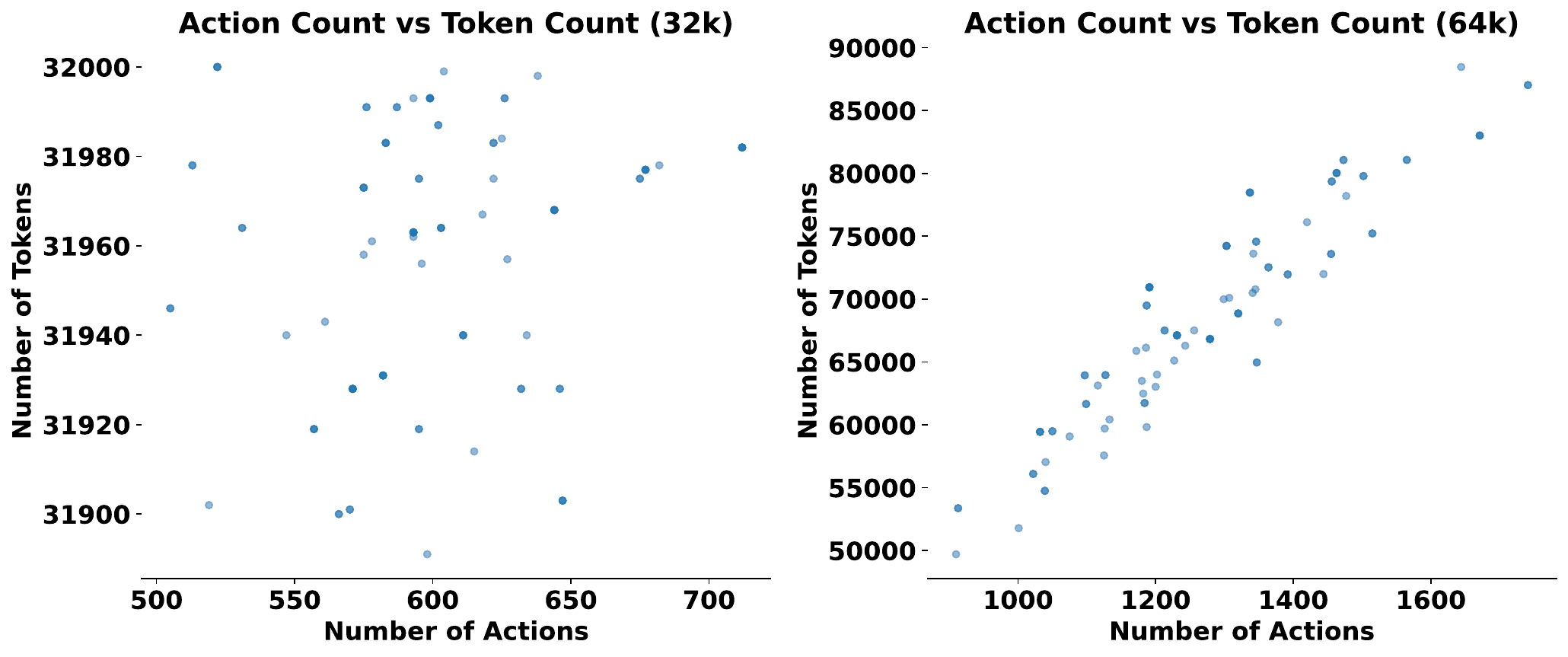}
    \caption{
    action token scatter for 32k and 64k setting.
    }
    \label{fig:action_token_scatter}
\end{figure}

\section{Full Dataset List and Construction Details}\label{app:dataset_summary}
Table~\ref{tab:all_datasets} lists all datasets used in this work, including the ShopTrajQA artifacts (summarized in the main text in Table~\ref{tab:dataset_summary}) and the additional transfer-evaluation benchmarks used to assess generalization. The shopping trajectories in ShopTrajQA (32k) are obtained by truncating the trajectories from the ShopTrajQA (64k) version. However, the retrieval QA pairs described in Section~\ref{sec:QA_Construction} are constructed separately. Since our data construction pipeline is easy to scale, we generate additional training data with questions that are different from those in the evaluation set, covering different trajectories and question. We then prompt a LLM (\texttt{Qwen3-Coder-30B}) to produce reference rollout (tool use trace) that (i) successfully execute all tool calls and (ii) obtain the correct final answer from the tool outputs. Compared with ShopTrajQA SFT, ShopTrajQA RL does not require predefined reference rollouts, as the model can discover correct rollouts through exploration during RL training.
\begin{table}[h]
\hspace{1mm}
\centering\scalebox{1}{
\begin{tabular}{lll}
\hline
Dataset    & Size  & Area               \\ \hline
\rowcolor{gray!12}
\multicolumn{3}{c}{\textbf{\textit{Training dataset}}} \\
ShopTrajQA SFT  & 762 & Shopping      \\
ShopTrajQA RL & 10822 & Shopping      \\
\midrule
\rowcolor{gray!12}
\multicolumn{3}{c}{\textbf{\textit{Evaluation dataset}}} \\
ShopTrajQA (32k)    & 100 & Shopping      \\
ShopTrajQA (64k)    & 100 & Shopping      \\
AIME 2024        & 30 & Mathematics                 \\
AIME 2025,   & 30 & Mathematics                 \\
2WikiMultihopQA     & 12576 &           Multi-hop QA       \\
HotpotQA &  7405  &   Multi-hop QA               \\
Bamboogle & 125 & Multi-hop QA \\
\hline
\end{tabular}}
\caption{Full list of datasets used in this work, including transfer-evaluation benchmarks.}
\label{tab:all_datasets}
\end{table}


\noindent{\bf Data Generation and Usage Statement.} We use Claude Haiku 4.5 \citep{claude_haiku45} to generate synthetic shopping trajectories that simulate user interaction environments. These trajectories provide contextual inputs for constructing our dataset.
The supervision signals, including reference answers and tool interaction traces used for supervised fine-tuning (SFT), are generated by \texttt{Qwen3-Coder-30B} \citep{yang2025qwen3}. Claude-generated trajectories are not used as supervision signals.
All generated data are used solely for research and benchmarking purposes.

\section{Data Quality Evaluation}\label{appendix:data_quality}

\noindent{\bf Experiment 1: Pairwise Trajectory Realism Evaluation with GPT-4.1}

We randomly sample trajectories from OPeRA and ShopTrajQA and construct 100 pairwise examples. In each pair, one trajectory is sampled from OPeRA and the other from ShopTrajQA. We store the source label for each trajectory as ground truth.
For each pair, we ask \texttt{GPT-4.1} to answer a multiple-choice question:
\begin{itemize}
\small
    \item The first trajectory is more like human-written.
    \item The second trajectory is more like human-written.
    \item Cannot distinguish.
\end{itemize}

The prompt template is shown below.

\begin{figure}[h]
\centering
\scalebox{0.8}{
\begin{tcolorbox}[title=Quality Evaluation Prompt]
You are evaluating the shopping behavior trajectories.

Decide which trajectory is more likely written/generated from real human behavior.

Be conservative and avoid overconfident guesses.

Return exactly one option token on the first line: FIRST, SECOND, or CANNOT\_DISTINGUISH.

Then on the second line, provide a short reason (<=40 words).

Trajectory FIRST:
\textcolor{blue}{first\_text}

Trajectory SECOND:
\textcolor{blue}{second\_text}

\end{tcolorbox}}
\end{figure}


We then compare \texttt{GPT-4.1} choices with the stored source labels. As shown in Table~\ref{tab:data_quality_pairwise}, GPT-4.1 selects the ShopTrajQA trajectory in 73 out of 100 cases, indicating that ShopTrajQA trajectories are generally judged as more human-like than OPeRA trajectories.

\begin{table}[H]
\small
\centering
\begin{tabular}{llll}
    & OPeRA & ShopTrajQA & Cannot distinguish \\ 
\hline
\#Case & 9     & 73         & 18             \\ 
\end{tabular}
\caption{Pairwise trajectory realism evaluation (100 pairs) using GPT-4.1.}
\label{tab:data_quality_pairwise}
\end{table}

\noindent{\bf Experiment 2: Training Data Comparison}

To further evaluate training-data effectiveness, we train the same models with two different data sources: DAPO-Math-17k \citep{yu2025dapo} and ShopTrajQA. Table~\ref{tab:model_comparison_mathdata} shows that models trained with ShopTrajQA consistently outperform those trained with DAPO-Math-17k on ShopTrajQA evaluation, indicating stronger transfer to shopping-domain long-context understanding.

\begin{table}[h]
\small
\centering
\scalebox{1}{
\begin{tabular}{lcc}
Model & 32k & 64k \\
\hline
Qwen3-1.7B-SFT-RLVR (DAPO-Math-17k) & 15 & 9 \\
Qwen3-1.7B-SFT-RLVR (ShopTrajQA) & \textbf{59} & \textbf{51} \\
Qwen3-4B-SFT-RLVR (DAPO-Math-17k) & 25 & 17 \\
Qwen3-4B-SFT-RLVR (ShopTrajQA) & \textbf{62} & \textbf{60} \\
\end{tabular}}
\caption{Training-data comparison on ShopTrajQA evaluation (accuracy, \%).}
\label{tab:model_comparison_mathdata}
\end{table}

\section{Additional Case Studies on Tool Use Failures}\label{app:case_studies}

This section presents two representative failure cases from \texttt{Qwen3- Coder-30B}. These cases illustrate why directly asking the model to write code over ultra-long shopping trajectories is error-prone, and why our tool design saves the trajectory as a local file and provides predefined utility functions and usage instructions.

\subsection{Case Study 1: Long-Context Code Generation Failure}

A straightforward approach is to place the entire shopping trajectory directly in the prompt and ask the model to write code to parse and analyze it. However, we found that this setup often leads to errors. One representative example is shown below.

\begin{figure}[h]
\centering
\scalebox{0.8}{
\begin{tcolorbox}[title=Case Study 1]
<code> python

history = """
[... Put all the shopping trajectory in the code ...]
"""

lines = history.strip().split('\textbackslash
')

first\_search = None

for line in lines:
    if "type [Search Amazon |" in line:
        first\_search = line
        break

first\_search\_index = lines.index ( first \_search)

click\_count = 0

for i in range(first\_search\_index):

~~    if "click [asin |" in lines[i]:
    
~~~~        click\_count += 1

print(click\_count)
``

</code>

<interpreter>
\textcolor{blue}{ValueError: None is not in list}
</interpreter>
\end{tcolorbox}}
\end{figure}

This case reveals several issues with directly embedding the full trajectory into the prompt and asking the model to code from scratch:

\begin{itemize}
    \item \textbf{Context omission risk.} When the model generates code that includes ultra-long context, parts of the trajectory may be omitted, truncated, or reformatted incorrectly.
    \item \textbf{Degraded coding reliability under long context.} The model must simultaneously preserve the raw context, understand the task, and write executable parsing logic. This increases the chance of implementation errors, as shown in Case 1.
    \item \textbf{High tool-use complexity.} The model is required to complete multiple steps at once: reading the trajectory, parsing the text, retrieving the relevant event, and writing correct code. This makes tool calling unnecessarily difficult.
\end{itemize}

This observation directly motivates our approach. Instead of asking the model to reconstruct the full context inside the generated code, we store the trajectory as a local file and provide predefined utility functions and tool-use instructions. This design reduces the burden of long-context handling and allows the model to focus on the reasoning step rather than low-level parsing.

Notably, in Table~\ref{tab:qwen3_results}, all settings with tool use are based on this design choice: the trajectory is saved locally, and the model is given explicit instructions and examples for using the provided tools.

\subsection{Case Study 2: Failure to Maintain Execution State Across Code Blocks}

Another issue arises even when the trajectory has already been wrapped into a local loading utility. In the following example, the model correctly loads the trajectory in one code block.

\begin{figure}[t]
\centering
\scalebox{0.8}{
\begin{tcolorbox}[title=Case Study 2]
<code> python

from data\_loader import load\_trajectory

conn, df = load\_trajectory('3')

df.head()

</code>

<interpreter>
Data preview:

row\_id            timestamp  ...                      search\_query

      0  2024-05-01 08:15:32  ...                  kids hair accessories

      1  2024-05-01 09:22:18  ...                      toddler headbands

      2  2024-05-01 10:05:44  ...         little girl headbands ages 2-6

      3  2024-05-01 10:18:27  ...                                  

      4  2024-05-01 10:22:15  ...                      power rangers dvd

[5 rows x 10 columns]

Columns:

Index(['row\_id', 'timestamp', 'action\_type', 'asin', 'product\_name', 'brand',

'color', 'product\_types', 'price', 'search\_query'],
dtype='object')

</interpreter>

<code>
df.columns
</code>

<interpreter>
\textcolor{blue}{NameError: name 'df' is not defined}
</interpreter>
\end{tcolorbox}}
\end{figure}

This case highlights a different failure mode: the model does not reliably account for execution-state boundaries across code blocks. In practice, the model may assume that previously defined variables remain available, while the actual execution environment requires the data to be reloaded in each new code block.

This issue suggests that successful tool use requires not only access to the right functions, but also awareness of execution state management. Without explicit guidance, the model may fail to reload the data before analysis, leading to avoidable errors such as \texttt{NameError}.

Our approach mitigates this issue by using RLVR training to encourage reloading the trajectory whenever a new execution block begins.


\section{Prompt for LLM}

\subsection{Prompt for LLM judge}\label{app:prompt_llmjudge}

\begin{figure}[t]
\centering
\scalebox{0.8}{
\begin{tcolorbox}[title=LLM Judge Prompt]

\textbf{You are an expert evaluator. Your task is to determine if a model's answer is factually correct for a given question.}

Question: \textcolor{blue}{question}

Expected Answer: \textcolor{blue}{reference\_answer}

Model Answer: \textcolor{blue}{model\_response}

- Compare the Model Answer with the Expected Answer.
- Minor differences in wording, formatting, or numeric representation are acceptable if the factual meaning is preserved.

- Product names (reasonably close brand/product names) are acceptable.

- Answer format flexibility: If the question asks for a product but the model gives the price, that's still correct if the price matches the expected product.

- If the Model Answer conveys the correct factual information (semantic equivalence), mark as CORRECT.

- If the Model Answer is factually incorrect, contradicts the Expected Answer, or contains only placeholders (e.g., "PLACEHOLDER\_FOR\_ANSWER", "\{\{final\_answer\}\}"), "None", or indicates no useful information (e.g., "No search event found"), mark as INCORRECT.

- Do NOT assume, invent, or speculate about missing information. Only evaluate the actual content provided in the Model Answer.

- Output ONLY one JSON object on the final line, and nothing else: \{\{"answer": "CORRECT"\}\} or \{\{"answer": "INCORRECT"\}\}.

\end{tcolorbox}}
\end{figure}

\subsection{Prompt for Shopping Trajectory Simulation}\label{appendix_shopping_prompt}


\begin{figure}[H]
\centering
\scalebox{0.8}{
\begin{tcolorbox}[title=Shopping Trajectory Simulation Prompt]

Given the input query data \{query\}, simulate a one month shopping trajectory for a single user. \\
Generate a shopping trajectory containing approximately:
search: 54.48\%, click: 35.31\%, add to Cart: 7.31\%, purchase: 2.90\%. \\

To ensure the global distribution is respected, the trajectory MUST enforce the following rule.

For every block of 100 actions:

- 55 search  - 35 click  - 7 add to Cart - 3 purchase actions.

These block-level ratios must be preserved across the entire trajectory to guarantee that

- search : click = 55 : 35 $\approx$ 1.5 : 1

- add to Cart : purchase = 7 : 3 $\approx$ 2 : 1.

Example timestamp format: 2024-05-01 19:30:25.

There are four possible action types: search, click, add to Cart, and purchase.

Each action must follow the exact formats below, with each action separated by a newline (\textbackslash n):

search: time-stamp search [Search Amazon $|$ \{query\_text\}].

click: time-stamp click [\{product\_id\} $|$ \{product\_title\}] (brand: \{product\_brand\}, color: \{product\_color\}, price: \{simulated\_price\}).

add to Cart: time-stamp add to Cart [\{product\_id\} $|$ \{product\_title\}] (brand: \{product\_brand\}, color: \{product\_color\}, price: \{simulated\_price\}).

purchase: time-stamp purchase [\{product\_id\} $|$ \{product\_title\}] (brand: \{product\_brand\}, color: \{product\_color\}, price: \{simulated\_price\}).

Each action begins with a timestamp and must satisfy the following constraints:

- add to Cart must always appear after a click on the same product.

- Purchase must always appear after its corresponding add to Cart on the same product.

- Before the user reaches the final accurate query \{search\}, simulate several exploratory searches (2 search actions) related to the underlying shopping intention.

Timestamps:

- Must be chronologically increasing 

- Spread naturally over May 2024.

Output format: \{``shopping\_trajectory'': ``action1\textbackslash naction2\textbackslash naction3\textbackslash n...''\}

Query: \textcolor{blue}{query}

\end{tcolorbox}}
\end{figure}

\section{Full Benchmark Results}\label{app:benchmark_table}

Table~\ref{tab:qwen3_results} reports the complete per-model benchmark accuracies with standard deviations that underlie Figure~\ref{fig:param_vs_perf}, for all commercial and open-source models evaluated without task-specific training. Within each block, models are ordered from larger to smaller scale; the parameter counts of the commercial models are undisclosed, so they are ordered by name only.

\begin{table}[h]
\centering
\scalebox{1}{
\begin{tabular}{lcc}
\toprule
Model & 32k & 64k \\
\midrule
\rowcolor{gray!12}
\multicolumn{3}{c}{\textbf{\textit{Commercial models}}} \\
gpt-5.5 (w/o tool) & 61.0 $\pm$ 1.1 & 46.0 $\pm$ 1.0 \\
gpt-5.5 (w/ tool) & \textbf{78.0 $\pm$ 1.3} & \textbf{73.0 $\pm$ 1.2} \\
gpt-5.6-sol (w/o tool) & 57.0 $\pm$ 1.0 & 44.0 $\pm$ 0.9 \\
gpt-5.6-sol (w/ tool) & 66.0 $\pm$ 1.1 & 63.0 $\pm$ 1.1 \\
Sonnet-4.5 (w/o tool) & 48.0 $\pm$ 0.9 & 38.0 $\pm$ 0.8 \\
Sonnet-4.5 (w/ tool) & 74.0 $\pm$ 1.2 & 66.0 $\pm$ 1.1 \\
Sonnet-4.6 (w/o tool) & 47.0 $\pm$ 0.9 & 40.0 $\pm$ 0.9 \\
Sonnet-4.6 (w/ tool) & 65.0 $\pm$ 1.1 & 59.0 $\pm$ 1.0 \\
\midrule
\rowcolor{gray!12}
\multicolumn{3}{c}{\textbf{\textit{Open-source models (large $\rightarrow$ small)}}} \\
gpt-oss-120B (w/o tool) & \textbf{62.6 $\pm$ 1.2} & \textbf{57.2 $\pm$ 1.0} \\
gpt-oss-120B (w/ tool) & 28.7 $\pm$ 0.9 & 35.5 $\pm$ 1.3 \\
Qwen3-Coder-30B (w/o tool) & 38.4 $\pm$ 1.1 & 24.7 $\pm$ 0.9 \\
Qwen3-Coder-30B (w/ tool) & 25.0 $\pm$ 0.8 & 22.4 $\pm$ 0.7 \\
gpt-oss-20B (w/o tool) & 22.3 $\pm$ 0.7 & 21.4 $\pm$ 0.8 \\
gpt-oss-20B (w/ tool) & 17.1 $\pm$ 0.6 & 18.8 $\pm$ 0.9 \\
Qwen3-4B (w/o tool) & 48.0 $\pm$ 1.0 & 47.6 $\pm$ 1.1 \\
Qwen3-4B (w/ tool) & 13.8 $\pm$ 0.4 & 9.3 $\pm$ 0.5 \\
Qwen3-1.7B (w/o tool) & 23.0 $\pm$ 0.6 & -- \\
Qwen3-1.7B (w/ tool) & 17.3 $\pm$ 0.5 & 14.7 $\pm$ 0.6 \\
\bottomrule
\end{tabular}}
\caption{Benchmark performance on ShopTrajQA under different context lengths, comparing commercial and open-source models without task-specific training. Results show accuracy (\%). Within each block, models are ordered from larger to smaller scale; parameter counts of the commercial models are undisclosed, so they are ordered by name only. ``--'' indicates that the prompt token length exceeds the maximum context window.}
\label{tab:qwen3_results}
\end{table}

\section{Training Progression of Our Models}\label{app:training_progress}

\begin{figure}[h]
\centering
\includegraphics[width=\columnwidth]{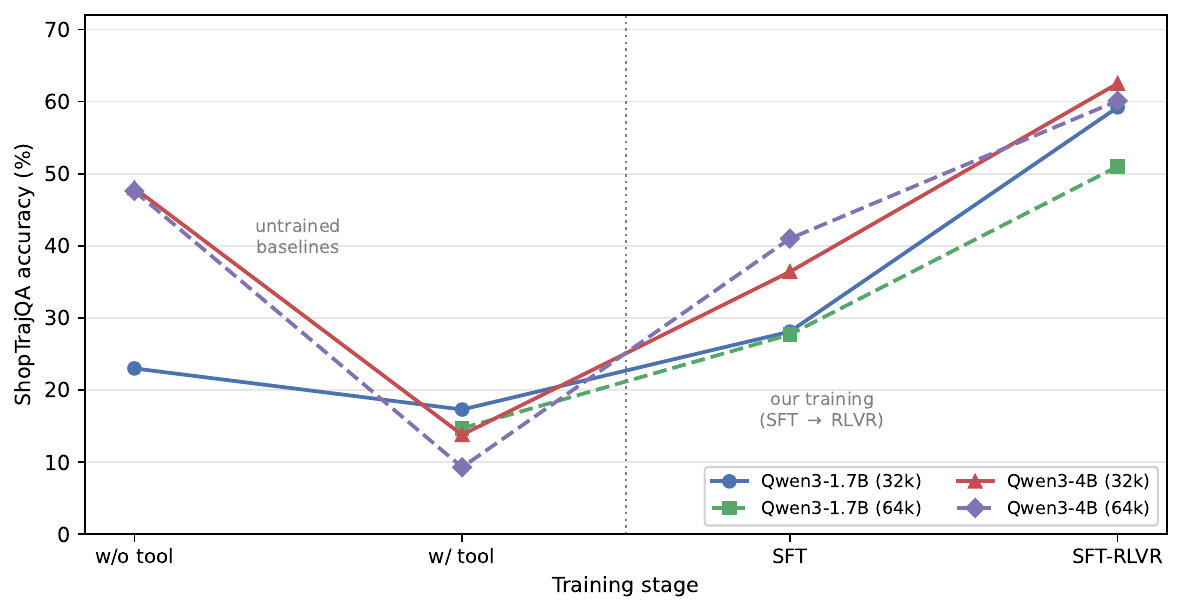}
\caption{Training progression on ShopTrajQA across stages (untrained w/o tool and w/ tool baselines, SFT cold-start, and final SFT-RLVR). Each line corresponds to one model $\times$ context setting (Qwen3-1.7B and Qwen3-4B, at 32k and 64k). The $y$-axis is accuracy (\%). The Qwen3-1.7B 64k line omits the pure-text (w/o tool) baseline, whose prompt exceeds the context window.}
\label{fig:training_progress}
\end{figure}

Figure~\ref{fig:training_progress} visualizes the training progression reported in Table~\ref{tab:training_results} for our two base models, Qwen3-1.7B and Qwen3-4B. Each line traces accuracy across the training stages (untrained w/o tool and w/ tool baselines, SFT cold-start, and final SFT-RLVR) for one model and context length. For both models, the untrained tool-calling accuracy is low, SFT recovers and surpasses the pure-text baseline, and SFT-RLVR delivers the largest gain. The Qwen3-1.7B 64k line omits the pure-text (w/o tool) baseline, whose prompt exceeds the model's context window.

\end{document}